# 3D shape retrieval basing on representatives of classes

M. Benjelloun, E. W. Dadi, and E. M. Daoudi

*Abstract*—In this paper, we present an improvement of our proposed technique for 3D shape retrieval in classified databases [2] which is based on representatives of classes. Instead of systematically matching the object-query with all 3D models of the database, our idea presented in [2] consist, for a classified database, to represent each class by one representative that is used to orient the retrieval process to the right class (the class excepted to contain 3D models similar to the query). In order to increase the chance to fall in the right class, our idea in this work is to represent each class by more than one representative. In this case, instead of using only one representative to decide which is the right class we use a set of representatives this will contribute certainly to improving the relevance of retrieval results. The obtained experimental results show that the relevance is significantly improved.

## I. Introduction

Thanks to the current digitizing and modeling technologies, the number of accessible and available 3D models, large databases included, is increasing. This has led to the development of 3D shape retrieval systems that, given a query object, retrieve similar 3D objects. The need for efficient methods in order to ease navigation into related large databases, and also to structure, organize and manage this new multimedia type of data, has become an active topic in various research communities such as computer vision, computer graphics, mechanical CAD, and pattern recognition.

Various 3D shape retrieval methods have been proposed in the literature [1,2,3,4,5,7,8]. All recent methods are based on shape indexing process which is consists to designing an efficient canonical characterization of the 3D shape of the model which is considered as its descriptor. Since the descriptor serves as a key in the search process, it is a critical kernel with a strong influence on the retrieval performances (i.e., computational efficiency and relevance of the results). A good 3D shape retrieval method must satisfy at least two conditions simultaneously [5]:

- The relevance: the top-k 3D objects returned by the method must be the most similar to the query;
- The speed up : the retrieval results should be fast.

In most of existing methods, it is unlikely to have the two conditions satisfied simultaneously. Moreover, for the most 3D shape retrieval approaches used in the literature, the matching is systematically performed with all 3D models in the database. Unfortunately, these approaches have several disadvantages:

- For the large database, the matching becomes increasingly difficult and needs more computational times; which does not allow the large scale retrieval.
- For the relevance of the results, the first retrieval results contain, in general, some objects that are not similar to the query.
- For the top-k answers, we have to wait until the matching will be completed with all the 3D models in the database, even if, only the first top k answers are needed.

To overcome disadvantages of the classical approaches we have proposed in a previous work [2] an efficient technique that restricts the shape matching process on a subset of "good candidates" by selecting the right 3D models that could be the best answers (the most similar) to a given 3D-object query. For a database classified into several classes, our idea is to orient the retrieval to the right class that contains similar objects to the query. In this case, the matching will not be done systematically with all objects within the database. The proposed solution is to represent each class by one representative which is a 3D-object selected among objects of this class, and then, according to the result of the matching between the query and the representatives, the retrieval will be oriented to the right class.

However, using only one representative to orient the retrieval may have the risk of not falling into the right class. In this work, we propose an improvement of our proposed technique [2] by representing each class by more than one representative. For a given class (human for example), the object can take different forms, situations and positions (for human : walking, sitting, sleeping), this why it is better to take into account all these situations, by representing each one by a representative. To do this, our solution in this paper consists of regrouping a set of subclasses of the same category (see Fig.1) as one class and represent it by a set of representatives each one corresponds to its subclass.

The retrieval process of our technique, using a set of representatives to represent each class, is performed as

E. W. Dadi is with National School of Applied Sciences, LaRi Laboratory, University of Mohammed First, MOROCCO (e.dadi@ump.ma).

M. Benjelloun is with Department of Computer Science Faculty of Engineering University of Mons, BELGIUM (e-mail: mohammed.benjelloun@umons.ac.be).

E. M. Daoudi is with Faculty of Sciences, LaRi Laboratory, University of Mohammed First, MOROCCO (m.daoudi@fso.ump.ma).

follows:
- First, for each subclass, we select the best representative using our proposed algorithm in [2].
- Next, we compare the query with all representatives.
- Then, for a query object, we select the right class that could contain the best-expected answers.
- Finally, the retrieval process will be launched in the selected right class.

The rest of the paper is organized as follows. In section 2 we present an overview of our proposed technique. Section 3 is devoted to the proposed improvement. The experimental results are presented in section 4. Section 5 provides some perspectives and concludes the paper.

| **Class 1 : *Airplanes*** Nbrs Of Objects(NOO): 98 | ***Sub-class 1*** Biplane NOO : 14 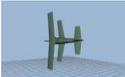 | ***Sub-class 2*** Commercial NOO: 11 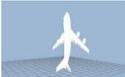 | ***Sub-class 3*** Fighter_jet NOO: 50 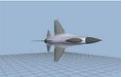 | ***Sub-class 4*** Glider_airplane NOO: 18 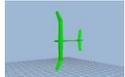 | ***Sub-class 5*** Stealth_bomber NOO: 5 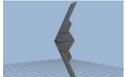 |
|---|---|---|---|---|---|
| **Class 2 : *Humans*** NOO : 78 | ***Sub-class 1*** bipeb NOO: 50 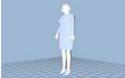 | ***Sub-class 2*** Arms_out NOO: 20 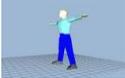 | ***Sub-class 3*** Walking NOO: 8 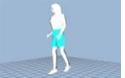 | | |
| **Class 3: *Quadruped*** NOO : 13 | ***Sub-class 1*** Dog NOO: 7 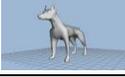 | ***Sub-class 2*** Horse NOO: 6 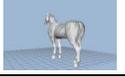 | | | |
| **Class 4: *Blades*** NOO : 26 | ***Sub-class 1*** Axe NOO: 4 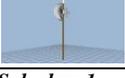 | ***Sub-class 2*** Knife_blade NOO: 7 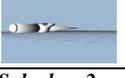 | ***Sub-class 3*** Sword_blade NOO: 15 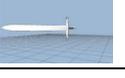 | | |
| **Class 5: *Chairs*** NOO : 26 | ***Sub-class 1*** Dining_chair NOO: 11 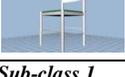 | ***Sub-class 2*** Desk_chair_seat NOO: 15 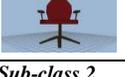 | | | |
| **Class 6: *Plants & trees*** NOO : 60 | ***Sub-class 1*** Bush_plant NOO: 9 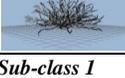 | ***Sub-class 2*** Flowers_plant NOO: 4 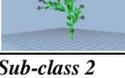 | ***Sub-class 3*** Potted_plant NOO: 26 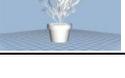 | ***Sub-class 4*** Barren_tree NOO: 11 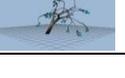 | ***Sub-class 5*** Conical_tree NOO: 10 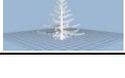 |
| **Class 7: *Cars*** NOO : 24 | ***Sub-class 1*** Race_car NOO: 14 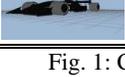 | ***Sub-class 2*** Sedan_car NOO: 10 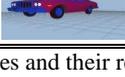 | | | |

Fig. 1: Classes and their representatives

## II. AN OVERVIEW OF OUR PROPOSED TECHNIQUE

In general, the 3D-shape retrieval process is performed in two stages: the first one consists in computing the descriptor of the query object and then, this descriptor is matched with the descriptor of each 3D model of the database. Note that, the query descriptor is computed online while the descriptors of the 3D models in the database are computed offline. The similarity between two descriptors is quantified by a dissimilarity measurement. For the standard retrieval approach proposed in the literature, the matching is systematically performed with all objects in the database; unfortunately, these approaches have several disadvantages:
- For large databases, the matching becomes increasingly difficult and needs more computational times, which does not permit large scale retrieval.
- The relevance of the results: in general, the expected answers are returned with some objects that are not similar to the query.
- Even if for the first top k answers, we have to wait until the matching with all the 3D models in the database is completed.

Our technique, proposed in [2], is based on the following idea: why the matching is systematically performed with all 3D models in the database and does not only be done with those are similar to the query. Assume that the database is classified into several classes (See Fig.1). The idea consists in representing each class of the database by one 3D object (the best one) selected among all 3D models of its class and then ordering the classes by order of similarity with the query by computing the distance between the representatives of classes and the query object. The retrieving process is primarily performed in the most similar class, and if necessary, it will be continued in the other classes, by order of priority. Our approach is performed in the following steps:
- Select representatives of classes.
- Ordering the classes by order of similarity. In particular, finding nearest class.
- Launching the 3D retrieving process in classes.

*A. Algorithm for selecting representatives*

A representative of a given class is a 3D object selected among all its 3D models. Our proposed algorithm for selecting representatives is based on the following idea: selecting the 3D object that is the closest to all 3D models of its class. The steps of this algorithm are given as follows:

| **Algorithm 1:** representative selection |
|---|
| For each 3D object k belonging the target class, do |
|    For each 3D object, i belonging the target class, do |
|       Compute the distance between object k with object i |
|    EndFor |
|    Compute the average distance of object k |
| EndFor |

*B. Retrieval process according to order of classes*

Representatives of classes serve as keys to guide and to order the retrieval into classes. This ordering is obtained by calculating the distance between the representative of each class and the query object. The class whose representative has the minimum distance with the query is considered as the right class; the class which contains 3D models most similar to the query. The process of the ordering is as follows:
- The query object is compared with the representative of each class by using a given 3D shape retrieval method, this last should be the most efficient in term of relevance even if it is not computational efficiency. The result of this step is a set of distances obtained performing this comparison.
- The classes are sorted according to the obtained distances. The class whose representative has the minimum distance is considered as expected class that contains the most similar objects to the query.

After the classes are sorted, the process of the retrieval is performed as follows:
- The retrieval will be started firstly in the nearest class (the obtained as the expected class) using a given 3D-object retrieval method. When the matching process is completely done with all objects in the class, the results can be returned.
- If the returned results are not satisfied, the retrieval process can be repeated recursively in the remaining classes according to the priority order until the results are satisfied or all classes are explored.

III. OUR IMPROVEMENT

In order to increase the chance to fall in the right class, our idea in this paper is to represent each class by more than one representative instead of what's proposed for the first version of our technique [2]. For a given class (human for example), the object can take different forms, situations and positions (for human : walking, sitting, sleeping), this why it is better to take into account all these situations, by representing each one by a representative. Note that as the principal objective of our improvement is to represent each class by more than one representative. For a given class, it is possible to choose as many representatives as we want. A possible solution consists of regrouping a set of subclasses of the same category (see Fig.1) as one class and represent it by a set of representatives each one is correspond to its subclass. Taking the example of Airplane class in the Fig.1, we have five subclasses of airplanes (biplane, commercial, fighter jet, glider, stealth bomber), each one is represented by a representative. Our solution is to regroup this subclasses as one class (Airplane class) and then represent it by the five representatives.

Our approach is performed in the following steps:
- For each subclass, we select the best representative using our proposed algorithm in [2].
- Each class will be represented by the set of representatives of its sub-class.
- For a given query, we compare it with all representatives. The class that one of its representatives has the minimum distance with the query is considered as the class which contains 3D object most similar to the query.

IV. EXPERIMENTAL RESULTS

During all steps of our approach, we have used the CM-BOF 3D retrieval method, proposed by Lian et al. [3] since it gives the best result comparing to several other methods, in particular, the view-based methods [3,5,7]. CM-BOF is 3D shape retrieval method, which uses Bag-of-Features and an efficient multi-view shape matching scheme. In this approach, a properly normalized object is first described by a set of depth-buffer views captured on the surrounding vertices of a given unit geodesic sphere. Then each view is represented as a word histogram generated by the vector quantization of the view's salient local features. The dissimilarity between two 3D models is measured by the minimum distance of their all (24) possible matching pairs.

We made our tests on the Test Princeton 3D Shape Benchmark database [6] (907 models categorized within 92 distinct classes).

For our test, we have chosen 22 classes considered as subclasses. The number of general classes obtained by regrouping 22 classes is 7 (Airplanes, Humans, Quadruped, Blade, Chair, Plant & tree and Cars). Fig.1 show the representative of each subclass selected using algorithm1, the name and the number of objects in each subclass.

### A. Precision of our retrieval technique

In Table 1 we report the precision of our retrieval technique for one representative (first version of our technique) and for a set of representatives (The new version with improvement).
- For the first version : In this case each subclass is represented by one representative.
- The new version :each classes is represented by a set of representatives.

For the experimental tests, we considered all 3D models of each classes as queries and then we try to reclassify them basing on representatives. The objective is to calculate the successful rate (SR) in each class which is defined as the number of queries of class i that are rightly oriented to the class i divided by the total number of objects of the class i. The idea is like doing a supervised classification. The steps of this experimentation is performed as follows :
- Each 3D-object of the database (the 3d models of each class) is considered as a query. In our case, we have 312 3D objects as query.
- Each query object is matched with the representatives of each class; in order to determine the right class.
- For each class i $(C_i)$, we compute the SR.

$$SR(C_i) = Q_i / N_i$$

Where $Q_i$ : is the number of 3D object that are good classed using our technique and $N_i$ : is the number of object in class $C_i$.

To compare between the two versions (the new and the previous) we calculate the SR over each general class. For the previous version the SR is calculated on each subclass considered as classes, to obtain SR of general class we compute the sum of SRs of its subclass the

TABLE I
THE OBTAINED SUCCESSFUL RATE

|  | First version (one representative) | New version with improvement |
|---|---|---|
| **Class 1:** *Airplanes* NOO : 98 | 71,43% (70/98) | 90,82% (89/98) |
| **Class 2 :** *Humans* NOO : 78 | 58,97% (46/78) | 91,03% (71/78) |
| **Class 3:** *Quadruped* NOO : 13 | 46,15 % (6/13) | 76,92% (10/13) |
| **Class 4:** *Blades* NOO : 26 | 53,85% (14/26) | 88,46% (23/26) |
| **Class 5:** *Chairs* NOO : 26 | 76,92% (20/26 ) | 80,77% (21/26) |
| **Class 6:***Plants & trees* NOO :60 | 43,33% (26/60) | 71,67% (43/60) |
| **Class 7:** *Cars* NOO : 24 | 91,67% (22/24) | 95,83% (23/24) |

The obtained results show that the use of a set representatives improves significantly the relevance which is logical because its permits to increase the chance to fall into the right class.

### V. CONCLUSION

In this paper we have proposed an improvement of our previous proposed technique. The idea is to represent each class by a set of representatives instead of only one representative in the previous version. Experimental results show that the relevance is significantly improved.